\documentclass[10pt,twocolumn,letterpaper]{article}

\usepackage{cvpr}              %
\usepackage{float}

\newcommand{\methodname}{TAPNext++}
\newcommand{\internalds}{$\text{DynHumans}$}
\newcommand{\dmin}{d_{\text{min}}} %
\newcommand{\ajrd}{\text{AJ}_{\mathrm{RD}}} %

\usepackage{pgf}
\usepackage{tikz}
\usepackage[table]{xcolor}

\usetikzlibrary{external}
\makeatletter
\DeclareRobustCommand\onedot{\futurelet\@let@token\@onedot}
\def\@onedot{\ifx\@let@token.\else.\null\fi\xspace}
\def\eg{e.g\onedot} 
\def\ie{i.e\onedot}

\def\etal{et~al\onedot}

\makeatother

\newcommand{\boldparagraph}[1]{\vspace{0.5em}\noindent{\bf #1.}}

\renewcommand{\paragraph}[1]{\boldparagraph{#1}}

\definecolor{darkgreen}{rgb}{0,0.7,0}
\definecolor{newyellow}{rgb}{1,0.8,0.05}
\definecolor{newgreen}{rgb}{0.2,0.8,0.2}

\definecolor{cvprblue}{rgb}{0.21,0.49,0.74}
\usepackage[pagebackref,breaklinks,colorlinks,allcolors=cvprblue]{hyperref}

\def\paperID{******} %
\def\confName{CVPR}
\def\confYear{2026}

\title{\methodname: What's \textit{Next} for Tracking Any Point (TAP)?}

\author{
Sebastian Jung$^{*,\text{\textdagger},3,4}$ \quad
Artem Zholus$^{*,5,7,8}$ \quad %
Martin Sundermeyer$^{1}$ \quad
Carl Doersch$^{2}$ \quad
Ross Goroshin$^{2,5,6}$ \\
David Joseph Tan$^{1}$ \quad
Sarath Chandar$^{5,7,8,9}$ \quad
Rudolph Triebel$^{3,4}$ \quad
Federico Tombari$^{1,10}$ \\
{\small $^1$Google \quad $^2$Google DeepMind \quad $^3$German Aerospace Center (DLR) \quad $^4$Karlsruhe Institute of Technology (KIT)}\\[-0.5ex]
{\small $^5$Mila - Quebec AI Institute \quad $^6$Université de Montréal \quad $^7$Chandar Research Lab}\\[-0.5ex]
{\small $^8$Polytechnique Montréal \quad $^9$Canada CIFAR AI Chair \quad $^{10}$Technical University Munich (TUM)}
}

\begin{document}

\twocolumn[{%
\renewcommand\twocolumn[1][]{#1}%
\maketitle
\vspace{-3.0em}
\begin{center}
    \includegraphics[width=1.0\linewidth]{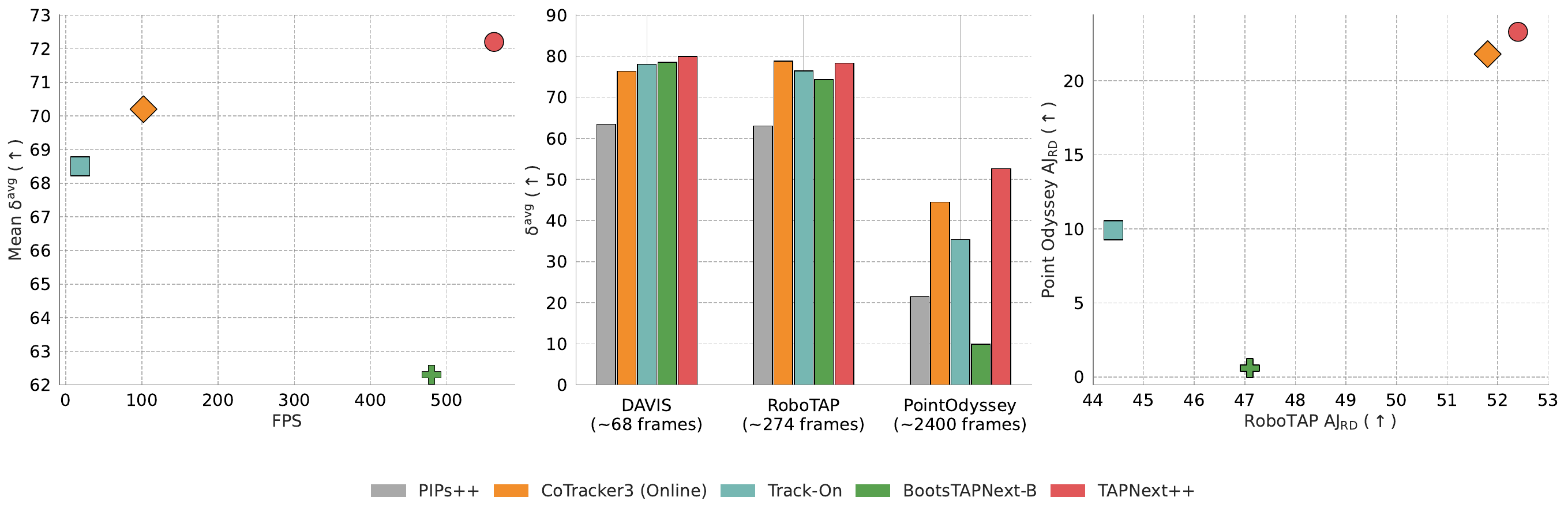}
    \vspace{-0.8cm}
    \captionof{figure}{\textbf{Overview.} \methodname~sets a new state-of-the-art for online point tracking by simultaneously improving accuracy, speed, and long-term robustness. \textit{Left:} Our method reaches the best speed-accuracy trade-off, achieving higher mean $\delta^{avg}$ than competing online methods while being substantially faster. \textit{Middle:} Despite running frame-by-frame without explicit memory, \methodname~outperforms previous methods on long-sequence benchmarks. \textit{Right:} \methodname~also demonstrates superior robustness to occlusion and re-appearences, achieving state-of-the-art $\ajrd$ scores, a new metric evaluating point tracking after re-detection, on both RoboTAP and PointOdyssey.}
    \label{fig:cover}
\end{center}
}]

\def\thefootnote{*}\footnotetext{Equal contribution.}\def\thefootnote{\arabic{footnote}}

\def\thefootnote{\textdagger}\footnotetext{Work done during an internship at Google.}\def\thefootnote{\arabic{footnote}}

\begin{abstract}
Tracking-Any-Point (TAP) models aim to track any point through a video which is a crucial task in AR/XR and robotics applications.  
The recently introduced TAPNext approach proposes an end-to-end, recurrent transformer architecture to track points frame-by-frame in a purely online fashion -- demonstrating competitive performance at minimal latency.
However, we show that TAPNext struggles with longer video sequences and also frequently fails to re-detect query points that reappear after being occluded or leaving the frame.
In this work, we present \methodname, a model that tracks points in sequences that are orders of magnitude longer while preserving the low memory and compute footprint of the architecture.
We train the recurrent video transformer using several data-driven solutions, including training on long 1024-frame sequences enabled by sequence parallelism techniques. 
We highlight that re-detection performance is a blind spot in the current literature and introduce a new metric, Re-Detection Average Jaccard ($\ajrd$), to explicitly evaluate tracking on re-appearing points. 
To improve re-detection of points, we introduce tailored geometric augmentations, such as periodic roll that simulates point re-entries, and supervising occluded points.
We demonstrate that recurrent transformers can be substantially improved for point tracking and set a new state-of-the-art on multiple benchmarks. Model and code can be found at \url{https://tap-next-plus-plus.github.io}.
\end{abstract}
    
\section{Introduction}
\label{sec:intro}

Tracking Any Point (TAP) in a video is a fundamental computer vision problem with many downstream applications in robotics, augmented reality, video editing, and 3D/4D reconstruction. 
Depending on the application, point tracking is performed online or offline, in 2D or 3D and under varying computational constraints. 
In this work, we focus on highly efficient, frame-by-frame 2D point tracking which is very relevant for AR applications such as anchoring digital content in the real world.
In contrast to optical flow estimation, which predicts dense pixel-level correspondences between adjacent frames, TAP methods attempt to consistently track points over multiple frames under strong appearance changes and occlusions. %
However, despite recent progress~\cite{Zholus_2025_ICCV,aydemir2025trackon2,Karaev_2025_ICCV}, state-of-the-art methods struggle to achieve robust performance in long sequences as shown in PointOdyssey~\cite{Zheng_2023_ICCV}. 
Common strategies to counteract this issue are: (1) tracking methods that operate over a sliding window of frames instead of one frame at a time~\cite{Karaev_2025_ICCV,harley2025alltracker}, (2) hard coding an explicit memory representation of past frames~\cite{aydemir2025trackon,aydemir2025trackon2}, (3) re-detecting points with appearance-based foundation models~\cite{aydemir2025trackon2} and (4) keeping the first frame as a persistent input at every time step~\cite{harley2025alltracker}. 
However, as our experiments demonstrate, these remedies fail to solve the root issue of temporally deteriorating representations and typically increase runtime and memory usage.
\begin{figure*}[t]
    \centering
    \includegraphics[width=2.0\columnwidth]{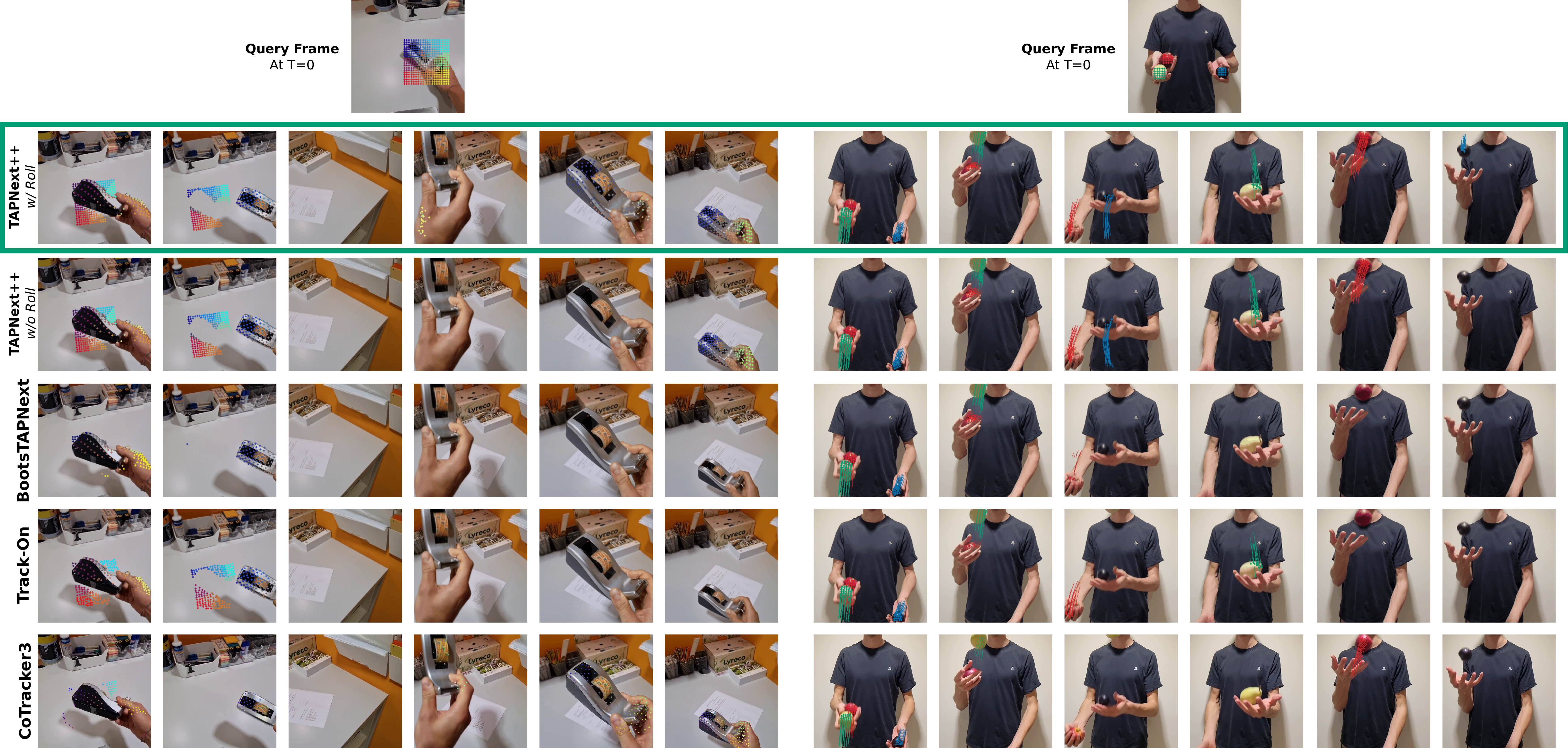}
 \caption{\textbf{Qualitative Comparison on Challenging Re-entry Scenarios.} We compare \methodname~ with prior methods on two sequences featuring point re-entries at varying locations. \textit{Left:} An object leaves the frame on the right and re-enters from the left. BootsTAPNext and Track-On fail to re-detect the object upon re-entry. In contrast, CoTracker3 and \methodname~ successfully relocate the points. However, CoTracker3 fails to track static points on the texture-less table, which \methodname~ tracks correctly. The variant \methodname~ without Roll also struggles with immediate re-detection, highlighting the importance of roll augmentation for re-entry robustness. \textit{Right:} Three objects are juggled, repeatedly leaving and re-entering the frame. Point trajectories for the last 8 frames are visualized as traces for clarity. \methodname~ with roll is the only method that robustly tracks all objects through these challenging dynamics, all competing methods fail.}
    \label{fig:qual_ex}
    \vspace{-0.5cm}
\end{figure*}

TAPNext~\cite{Zholus_2025_ICCV} reformulates point tracking as next token prediction and deploys linear recurrent SSM layers~\cite{patraucean2024trecvitrecurrentvideotransformer} to track scene dynamics, particularly of query points, over time. 
Its constant-sized, linear-recurrent memory allows TAPNext to jointly track 1024 points at 348 FPS on an H100 GPU, demonstrating the efficiency required for mobile applications in robotics and augmented reality.
However, as also noted in concurrent work~\cite{aydemir2025trackon2}, the BootsTAPNext model trained on a large number of short videos at a fixed $256\times256$ resolution~\cite{doersch2022tap} fails to generalize to longer sequences~\cite{Zheng_2023_ICCV}. 
In this work, we present \methodname, a model demonstrating that these tracking failures are not inherent to the underlying recurrent video transformer architecture and can be effectively tackled through improved training strategies and datasets.

Tracking points over long time horizons reveals another severe limitation of current TAP methods: when points reappear after being occluded or leaving the view, their re-detection frequently fails. This limits applications, \eg in AR/VR, where both the camera and the scene are dynamic and reappearances are a frequent occurrence. Recent approaches resort to hard-coded appearance-based re-detection without relying on motion priors required to disambiguate, \eg,~multiple instances of visually similar points. Some approaches~\cite{Karaev_2025_ICCV} track occluded points explicitly within the image, which improves re-detection after short-term occlusions, but usually fail for long-term occlusions, particularly when tracked points venture outside the image boundaries. Other methods~\cite{Xiao_2025_ICCV} attempt to build an explicit 3D map using monocular video depth foundation models~\cite{li2025megasam}, assuming a mostly static scene and generous computational resources.

We find that the re-detection limitation is a blind spot in current point tracking metrics and evaluations. The closest related metric is the survival rate~\cite{Zheng_2023_ICCV}, which measures the average number of frames until point tracking failure. However, the survival rate does not directly measure re-detection performance because it includes many points that are always visible. Therefore, we propose a variation of the Average Jaccard metric specifically targeting the performance of point tracking after reappearance and evaluate recent approaches on it. 
\\
\\
Our main contributions are as follows:
\begin{itemize}
    \item We introduce methods that allow scaling the training of TAPNext architectures to long sequences and multiple resolutions -- resulting in a lightweight, fully online model that sets a new state-of-the-art on multiple point tracking benchmarks.
    \item We introduce Kubric-1024, which extends the original synthetic Kubric dataset~\cite{greff2022kubric} to 1024 frames and analyze the effects of long-sequence training on memory.
    \item We propose novel re-detection metrics that directly measure the accuracy of point tracks after re-appearance and improve on these metrics through tailored geometric augmentations.
   \item We demonstrate how our contributions effectively enhance the memory of the TAPNext model, outperforming all existing methods in long-context and re-detection scenarios.
    We highlight that we \textit{do not} require any new model architectures to surpass existing methods while further enhancing the compute and memory efficiency of the original TAPNext. 
\end{itemize}

\section{Related Work}
\label{sec:related_work}

\paragraph{Point Tracking} Tracking Any Point (TAP)~\citep{doersch2022tap} is a computer vision task where the model is tasked to find the position and visibility of a given \textit{query} point throughout a video. Here, we focus on sequential 2D point tracking, without access to future frames, for its relevance in compute-constrained mobile applications. 

\paragraph{Windowed Methods} The majority of prior methods for TAP perform point tracking in two stages. First, they precompute image or video features and then they use these features in an appearance matching and iterative refinement process to track predictions over a window of frames or an entire video. More specifically, TAPNet \citep{doersch2022tap} performs tracking via a simple feature matching on CNN features. TAPIR \citep{Doersch_2023_ICCV} introduces an iterative refinement procedure within TAPNet. BootsTAPIR \citep{Doersch_2024_ACCV} extends the method with a large-scale, real world dataset to bootstrap self-supervised training on natural, unlabeled videos. The TAPTR family \citep{li2024taptr,li2024taptrv2,qu2024taptrv3} performs tracking using a DETR-like architecture. LocoTrack \citep{cho2024local} incorporates bidirectional 4D correspondences for fast and accurate tracking. CoTracker~\cite{karaev2024cotracker} presents a hybrid CNN-transformer architecture demonstrating the benefit of tracking all point queries jointly. CoTracker3 \citep{Karaev_2025_ICCV} extends CoTracker by performing semi-supervised finetuning on a real-world pseudo-labeled dataset, showing the importance of training data for point tracking. AllTracker~\citep{harley2025alltracker} incorporates high-resolution, pixel-dense point tracking by a multi-view optical flow optimization while re-inferring the query frame to avoid temporal degradation. However, on long sequences as present in the PointOdyssey~\cite{Zheng_2023_ICCV} benchmark, points deviate significantly from their initialization, and attending on overlapping sliding windows or query frames is often insufficient, as shown in Fig.~\ref{fig:cover}. 

\paragraph{Frame-by-frame Methods} Arguably, defining such task specific context windows could be avoided if temporal information would simply propagate from frame to frame. The TrackOn~\citep{aydemir2025trackon,aydemir2025trackon2} family performs online, frame-by-frame tracking using patch classification and refinement to identify correspondences and propagates temporal information through explicit memory banks. Notably, TAPNext \citep{Zholus_2025_ICCV} shows that it is possible to efficiently perform frame-by-frame TAP without explicit memory banks and without point tracking-specific inductive biases. Instead, TAPNext adopts a token-based modeling formulation and casts TAP as temporal masked decoding. TAPNext uses an architecture~\citep{patraucean2024trecvitrecurrentvideotransformer} with alternating ViT and State-Space-Model (SSM) blocks for spatial and temporal feature propagation, respectively. The linear recurrent layers are efficient to train and infer on modern GPU hardware allowing state-of-the-art processing speed for point tracking. However, despite performing well on common metrics and short-term benchmarks, we demonstrate its severe limitations in propagating information over long time horizons and in re-detecting points after longer occlusions, two crucial capabilities for real-world applications. In this paper, we overcome the limitations of TAPNext without relying on external appearance-based foundation features~\cite{oquab2023dinov2} that do not model motion.

\begin{figure}
    \centering
    \includegraphics[width=\columnwidth]{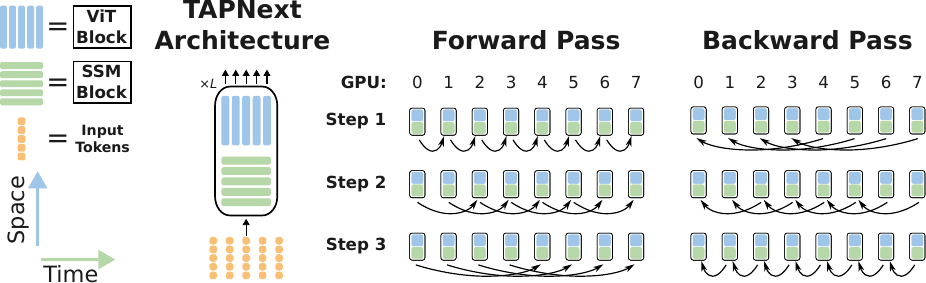}
    \caption{\textbf{Distributed Parallel Scan during Training.} On the left, the TAPNext architecture with its SSM and ViT Blocks is shown. Right, example of information flow during multi-gpu training using a distributed parallel scan. Note that information is only passed between the SSM Blocks of the GPUs and not the ViT Blocks. Only inter-GPU communication is shown, intra-GPU parallel scans are not shown for simplicity. For a detailed figure of the TAPNext architecture we refer to~\cite{Zholus_2025_ICCV}.}
    \label{fig:long_sequence_training}
    \vspace{-0.5cm}
\end{figure}

\section{Method}

\methodname, as TAPNext~\cite{Zholus_2025_ICCV}, processes videos in a causal online fashion. It processes videos of $T$ RGB frames, each of size $H\times W$ pixels and $Q$ point queries per video, each of which is an index of the form $(t,x,y)$. For each frame in a video, we apply a standard ViT-style linear projection followed by learned positional encodings that are added to the linear projections. As a result, we get $T\times h \times w$ visual tokens, where $h$ and $w$ are sizes after patchification. We apply spatial positional encoding for each point query. Specifically, the $(x,y)$ component serves as a 2D index in the positional embedding tensor. For each point query, we initialize $T$ \textit{track tokens}, where $T-1$ are filled with a \texttt{[MASK]} token and one, at position $t$ (from the point query), is filled with the position embedding of the given query representing the spatial coordinates $x$ and $y$. Consequently, we have $T\times Q$ point track tokens. We simply concatenate $T\times Q$ track tokens with $T\times h \times w$ visual tokens after position embedding over the spatial axis resulting in  $T\times ( h\times w + Q)$ spatiotemporal tokens. The TAPNext architecture performs attention only over the spatial dimension for every time step while causally propagating information for each token sequence independently in the temporal dimension using an SSM layer, see \cref{fig:long_sequence_training}.

While Zholus \etal~\cite{Zholus_2025_ICCV} showed that this architecture can be applied to standard point tracking benchmarks, our focus lies on identifying its limitations which we elucidate and address. One major flaw of the original TAPNext is its inability to track points for an extended temporal horizon ($\sim$150 frames). This is perhaps unsurprising: the model is trained on 48-frame sequences, meaning that longer sequences are out-of-domain. 
Although recent works~\cite{de2024griffin, patraucean2024trecvitrecurrentvideotransformer} show that the Linear-Recurrent-Unit used in the TAPNext architecture is able to perform well on sequences that are larger than the training sequences, these works train on longer sequences initially in order to learn stable update rules. Motivated by these results, we hypothesise that training on longer sequences also increases long-term tracking stability. 
Therefore, our first goal is to fine-tune a previously released TAPNext checkpoint on 1024 frames. 

\begin{figure}[t]
    \centering
    \includegraphics[width=1\columnwidth]{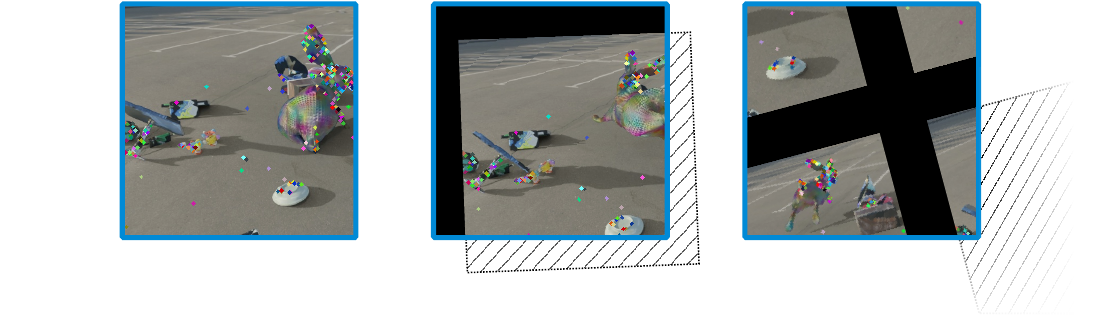}
    \vspace{-0.8cm}\caption{\textbf{Roll Augmentation.} Videos are rotated and translated smoothly over time, wrapping around the image with a gap. Ground-truth points are visualized.}
    \label{fig:roll_aug}
    \vspace{-0.5cm}
\end{figure}

\paragraph{Long-Sequence Training} Training TAPNext on long sequences requires fitting lengthy computation graphs and activations into memory, which exceeds single-GPU limits for sequences of 1024 frames. To enable end-to-end training on such sequences we employ sequence parallelism, sharding the input sequence only along the time dimension and assigning each chunk to a different GPU. TAPNext's architecture builds upon Real-Gated Linear Recurrent Units (RG-LRU), a variant of State Space Models (SSMs). Thanks to their linearity, SSM  temporal processing can be parallelized using the associative scan primitive. To exploit this property under sequence parallelism, we introduced a distributed parallel scan for both RG-LRU and temporal convolution layers during forward and backward passes. In this setup, each GPU first performs a local scan over its assigned sequence chunk. This is followed by an efficient, logarithmic-time merge operation across GPUs to combine hidden states over chunk boundaries, and a final local update based on the merged states. This distributed scan enables end-to-end training on long sequences by parallelizing temporal computation across multiple devices, overcoming single-device memory limitations while maintaining full temporal context. As can be seen in \cref{fig:long_sequence_training}, distributed parallel scan allows performing forward and backward pass in three instead of seven steps on eight GPUs compared to a na\"ive sequential communication.

\begin{figure}[t]
    \centering
    \includegraphics[width=1.0\columnwidth]{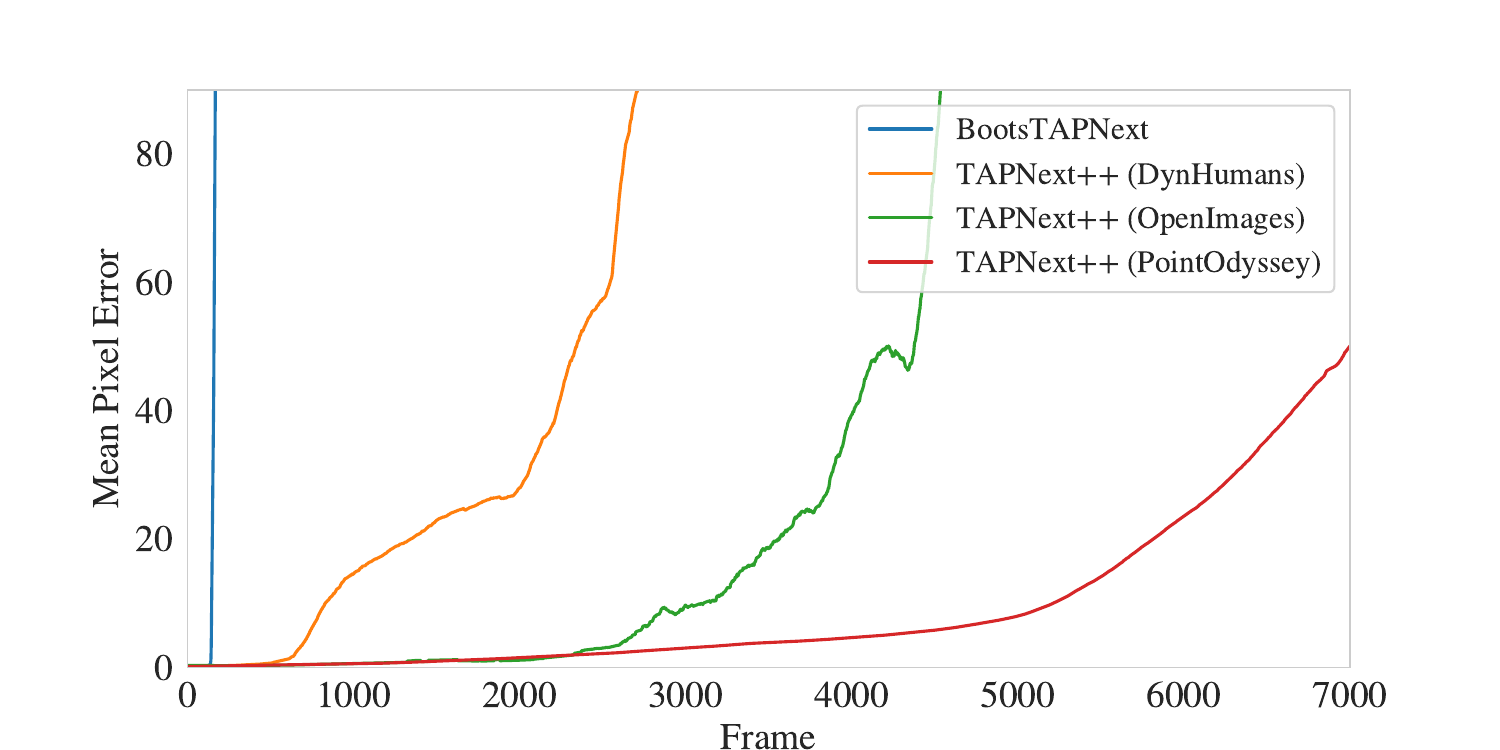}
    \caption{\textbf{Long-Sequence Stability.} Mean Pixel Error of 16x16 points after repeatedly inferring a static image through TAPNext variants. Finetuning on 1024-frame videos from PointOdyssey most effectively extends the memory longevity.}
    \label{fig:longevity}
    \vspace{-0.4cm}
\end{figure}

\begin{figure}[t]
    \centering
    \includegraphics[width=1\columnwidth]{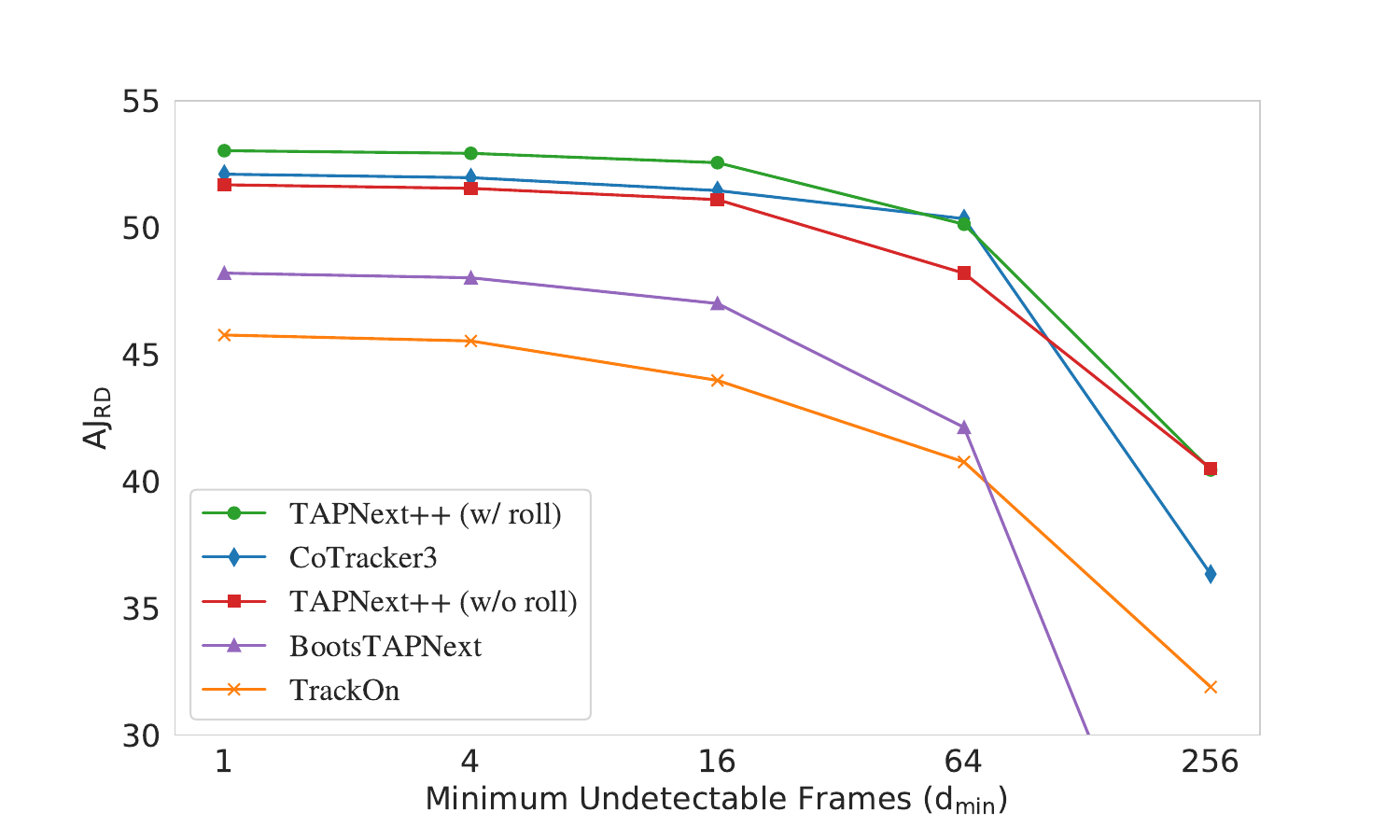}
    \caption{\textbf{$\ajrd$ Comparison.} Re-Detection Average Jaccard ($\ajrd$) for different $\dmin$ values on RoboTAP. TAPNext++ with roll augmentations preserves AJ after long periods of undetectability. The window-based CoTracker3 exhibits a drop in $\ajrd$ following a sequence of 256 undetectable frames.}
    \label{fig:aj_rd_plot}
    \vspace{-0.5cm}
\end{figure}

\begin{figure*}[t]
    \centering
    \includegraphics[width=2.0\columnwidth]{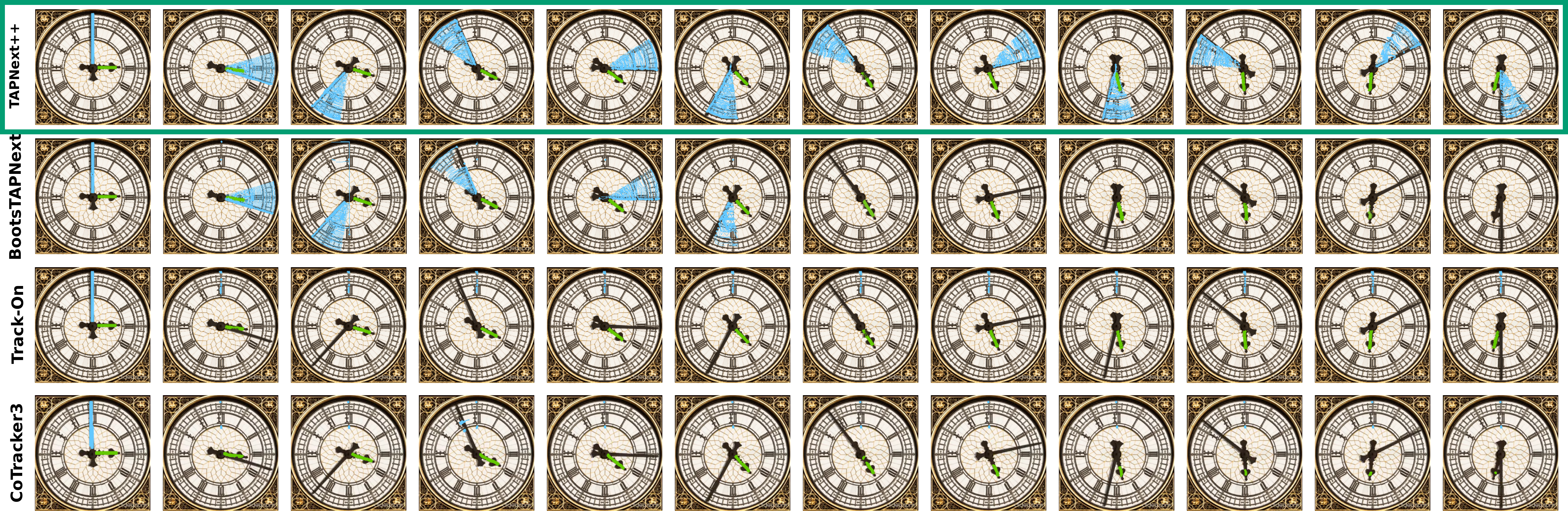}
    \caption{\textbf{Long-Term Dynamic Tracking Example.} We compare state-of-the-art online trackers on tracking clock hands, visualizing a trace of the last 6 frames. While the original BootsTAPNext, Track-On and CoTracker3 fail at tracking both clock hands for the full duration of the video, \methodname~tracks both until the end. CoTracker3 and Track-On are only able to track the hour hand due to its fine structure, additionally CoTracker3 loses the hour hand at the end of the video.}
    \label{fig:clock}
    \vspace{-0.5cm}
\end{figure*}

\paragraph{Long Sequence Data} Point tracking research requiring long sequence training data with ground truth annotations is currently limited mainly to PointOdyssey~\cite{Zheng_2023_ICCV}. Given that the training split of PointOdyssey (v1.2) contains only 131 videos, we identified a need for a new, large-scale, long-sequence tracking dataset. To address this, we generated a synthetic dataset using Kubric~\cite{greff2022kubric}, consisting of 10,000 videos with 1024 frames. Each scene uses an HDRI from the public Polyhaven~\cite{zaal2021polyhaven} library for realistic background and illumination. We populate scenes with 10--20 static and 1--10 dynamic objects sampled from GSO~\cite{downs2022google}, which are randomly placed via an overlap-avoiding algorithm within designated spawn regions. The camera follows a smooth trajectory between a random start and end position, sampled within a spherical shell around the world origin to ensure objects remain in view. To simulate more natural viewing patterns, this base path is augmented with sinusoidal noise, mimicking unsteady camera motion. Furthermore, to create more dynamic pans, the camera's look-at point is animated with sinusoidal noise and projected onto the scene's ground plane, rather than being fixed at the origin. To maintain object motion and interaction throughout the long sequences, we use the PyBullet~\cite{coumans2021} physics simulator and introduce velocity "bumps": if a dynamic object's velocity falls below a given threshold, a new random linear and angular velocity is applied. This velocity includes a slight bias toward the scene origin, encouraging objects to move within the camera's field of view and preventing scenes from becoming static over time.

\begin{table*}[h!]
    \caption{\textbf{Online Tracking Comparison.} We report results on multiple point tracking benchmarks and compare against current state-of-the-art online trackers.}
    \label{tab:benchmark_results}
    \centering
    \resizebox{\linewidth}{!}{
    \begin{tabular}{c l c c c c c c c c c c c c c c c c c c c}
    \toprule
    \textbf{\#}& \textbf{Method} & \textbf{Res.} & \multicolumn{4}{c}{\textbf{PointOdyssey}} & \multicolumn{3}{c}{\textbf{DAVIS}} & \multicolumn{3}{c}{\textbf{RGB-Stacking}} & \multicolumn{4}{c}{\textbf{RoboTAP}} & \multicolumn{3}{c}{\textbf{Kinetics}} & \multicolumn{1}{c}{\textbf{Mean}} \\
    \cmidrule(lr){4-7} \cmidrule(lr){8-10} \cmidrule(lr){11-13} \cmidrule(lr){14-17} \cmidrule(lr){18-20} \cmidrule(lr){21-21}
    & & & $\delta^{avg}$ $\uparrow$ &Survival $\uparrow$ & MTE $\downarrow$ & $\ajrd$ $\uparrow$ & AJ $\uparrow$ & $\delta^{avg}$ $\uparrow$ & OA $\uparrow$ & AJ $\uparrow$ & $\delta^{avg}$ $\uparrow$ & OA $\uparrow$ & AJ $\uparrow$ & $\delta^{avg}$ $\uparrow$ & OA $\uparrow$ & $\ajrd$ $\uparrow$ & AJ $\uparrow$ & $\delta^{avg}$ $\uparrow$ & OA $\uparrow$ & $\delta^{avg}$ $\uparrow$ \\
    \midrule
    1 & PIPs++ & $256\times256$ & 21.5 & 38.1 & 46.1 & -- & -- & 63.4 & -- & -- & 58.5 & -- & -- & 63.0 & -- & -- & -- & -- & -- & 51.6 \\
    2 & CoTracker2 & $384\times512$ & 30.2 & 55.2 & 215.0 & 3.9 & 62.2 & 75.7 & 89.3 & 67.4 &  78.9 & 85.2 & 58.6 & 70.6 & 87.0 & 33.3 & 48.8 & 64.5 & 85.8 & 64.0 \\
    3 & CoTracker3 (Online) &  $384\times512$ & \underline{44.5} & \underline{56.3} & \underline{20.7} & \underline{21.8} & 63.8 & 76.3 & 90.2 & \underline{71.7} & 83.6 & 90.2  & \textbf{66.4} & \textbf{78.8} & \textbf{90.8} & \underline{51.8} & \underline{55.8} & 68.5 & \underline{88.3} & \underline{70.2} \\
    4 & Track-On &  $384\times512$ & 35.4 & 47.5 & 33.5 & 9.9 & 65.0 & 78.0 & 90.8 & 71.4 &  \textbf{85.2} & \underline{91.7} & \underline{63.5}  & \underline{76.4} & 89.4 & 44.4 & 53.9 & 67.3 & 87.8 & 68.5 \\
    \midrule 
    5 & BootsTAPNext-B & $256\times256$ & 9.9 & 13.0 & 92.1 & 0.6 & \underline{65.2} & \underline{78.5} & \underline{91.2} & 66.2 & 78.3 & 86.8 & 62.6 & 74.3 & 88.4 & 47.1 & \textbf{57.3} & \textbf{70.6} & 87.4 & 62.3 \\
    6 & \textbf{\methodname} & $256\times256$ & \textbf{52.6} & \textbf{67.9} & \textbf{13.4} & \textbf{23.3} & \textbf{66.6} & \textbf{79.9} & \textbf{92.1} & \textbf{73.4} & \underline{84.8} & \textbf{95.1} & 61.1 & 75.2 & \underline{89.6} & \textbf{52.4} & 54.4 & \underline{68.7} & \textbf{89.0} & \textbf{72.2} \\
    
    \bottomrule
    \end{tabular}
    }
    \vspace{-0.5cm}
\end{table*}

\paragraph{Augmentations} Long-sequence tracking demands robustness to points that become invisible for extended periods, either through occlusion or by exiting and re-entering the camera's field of view. These re-entry events are common in real-world applications, such as augmented reality, but pose a significant persistent memory challenge for tracking models. This is especially challenging if points reappear far from their last known position, particularly for trackers that explicitly enforce local search windows. 
To improve the model's ability to handle such cases, we employ data augmentations that simulate camera jitter and roll during training. We apply temporally-smooth, sinusoidal random translations and rotations to each video sequence and its corresponding point tracks. Crucially, translations are implemented as periodic shifts; pixels and track coordinates shifted off one image boundary wrap around to the opposite side with a given margin. This mechanism explicitly simulates query points leaving the frame on one side and re-entering from another, forcing the model to rely more on appearance matching for re-detection, rather than assuming spatio-temporal proximity to the point's last visible position. To prevent double-appearance of the same tracking point in a single frame, we add a margin of $\sqrt{\left(\frac{H}{2}\right)^2+\left(\frac{W}{2}\right)^2}$ between the wrapped images.
This synthesis of challenging re-entry scenarios improves model robustness against spatial displacements in point trajectories. An example of the roll augmentation can be found in \cref{fig:roll_aug}.
Additionally we apply aspect ratio augmentations by picking a random aspect ratio between $0.5$ and $2.0$ during training.

\paragraph{Tracking Occluded Points via Weighted Loss} The standard loss function used in TAPNext only supervises coordinate prediction for points when they are visible. This provides no incentive for the model to predict plausible locations for points that are temporarily occluded, which can hinder their re-detection upon reappearance and negatively impact long-term tracking performance. To improve tracking robustness through occlusions, we modify the training loss to include supervision for points that are occluded but remain within the image boundaries, weighting the position loss of occluded points by $0.2$ compared to the position loss on visible points.

\paragraph{High-Resolution Fine-Tuning} Following prior work~\cite{karaev2024cotracker}, which shows high-resolution inputs improve point tracking, we fine-tune our best \methodname~checkpoint at $512 \times 512$. Since the TAPNext encoder uses a fixed $8 \times 8$ patch size, the larger input produces an expanded spatial token grid. Consequently, we resize the learned 2D positional embeddings from the base $32 \times 32$ grid to $64 \times 64$ using bicubic interpolation. The output layer size (256 + 256 bins) remains unchanged. The model is then fine-tuned on 256-frame sequences for $120\text{k}$ steps.

\section{Experiments}
\begin{table}[t!]
    \caption{\textbf{Inference Speed.} Speed comparison of \methodname~ to TAPNext, LocoTrack-B ($256\times256$), online CoTracker3 and Track-On running on Nvidia H100 GPUs. The latency metric is defined as the maximum (worst case) time between passing a frame to the model and receiving predicted points and it includes the time it takes to fill and process the initial frame buffer. \textit{(frame)} indicates per frame inference,  \textit{(window)} is when we track with non-overlapping chunks of 32 frames. All models are implemented in PyTorch with TAPNext++ relying on FlashAttention3~\cite{shah2024flashattention}.}
    \label{tab:speed}
    \centering
    \resizebox{\linewidth}{!}{
    \begin{tabular}{c|l|c|c}
    Query Points & \multicolumn{1}{c|}{Model} & Average FPS $\uparrow$ & Latency (ms) $\downarrow$ \\
    \midrule
    & LocoTrack-B (window) & 452 & $2210$  \\
    & CoTracker3 (online) & 102  & $80$   \\
    & Track-On (frame) & 28 & $36$ \\
256  & TAPNext (frame) & $189$  & $5.29$\\
    & \textbf{TAPNext++ (frame)} & $193$  & $\mathbf{5.18}$\\
    & TAPNext (window) & $480$  & $66$ \\
    & \textbf{TAPNext++ (window)} & $\textbf{562}$  & $57$ \\
    \hline
    & LocoTrack-B (window) & 124 & $8000$  \\
 & CoTracker3 (online) & 45  & $177$  \\
 & Track-On (frame) & 23 & $44$ \\
1024    & TAPNext (frame) & $182$  & $5.47$   \\
    & \textbf{TAPNext++ (frame)} & $191$  & $\mathbf{5.23}$\\
    & TAPNext (window) & $300$ & $106$ \\
    & \textbf{TAPNext++ (window)} & $\textbf{348}$  & $57$ \\
    \hline
    \end{tabular}
    }
\end{table}

\subsection{Metrics} We evaluate performance using the standard metrics from TAP-Vid~\cite{doersch2022tap}: Occlusion Accuracy (OA), positional error ($\delta^{avg}$), and Average Jaccard (AJ) which evaluates both occlusion and position accuracy. Additionally, for the PointOdyssey dataset, we report Survival Rate and Median Translation Error (MTE) as defined by Zheng~\etal~\cite{Zheng_2023_ICCV}.

While informative, these metrics average performance over entire tracks and do not specifically quantify a tracker's ability to re-detect points that reappear after being undetectable (i.e. occluded or out-of-frame) for many frames. 
\subsection*{Re-Detection Average Jaccard $\ajrd$}
To address this, we propose the \textit{Re-Detection Average Jaccard} ($\ajrd$), a new metric that measures tracking quality after a point reappears, conditioned on how long it was undetectable. We start with definitions:

\noindent\textbf{Undetectable Point:} A point is undetectable in frame $t$ if its ground-truth visibility $v_{t}$ is 0, either because it is occluded by another object or itself, or it is outside the image bounds. Otherwise, the point is detectable ($v_{t}=1$).

\noindent\textbf{Reappearance Event:} A reappearance event occurs for a track at frame $t_r$ if the point is detectable at frame $t_r$ but was undetectable for the $d$ preceding frames, \ie, $v_{t_r}=1$ and $v_{t_{r-1}}=\dots=v_{t_{r-d}}=0$.

\noindent\textbf{Duration of Undetectability:} The value $d \geq 1$ is the duration of undetectability.

\noindent\textbf{Eligible Reappearance Event:} To focus on how well trackers handle different lengths of time intervals for point disappearance, we only consider a reappearance event $i$ if its duration $d_i$ is \textit{strictly greater than the maximum disappearance duration of all previous reappearance events} for that same track. We call such events eligible. This ensures that for any given track, only events that set a new record for invisibility duration are counted. For example, if for a particular track the sequence of $d_i=\mathbf 1, 1, \mathbf 4, 3, 2$ then the \emph{eligible} reappearance events are highlighted in bold.\\
\\
A tracker might locate a point in frame $t_r$ but fail to track it in frames $t_r+1, t_r+2, \dots$, even if the point remains visible. $\ajrd$ is designed to measure this post-reappearance tracking quality, particularly for points that were undetectable for a long time.
We define $\ajrd^{\dmin}$ as follows:
First, identify all \textit{eligible} reappearance events where the duration of undetectability $d$ is greater than or equal to $\dmin$. For each such event occurring at a frame $t_r$, consider the track segment consisting of all frames from $t_r$ to the end of the video sequence. $\ajrd^{\dmin}$ is the Average Jaccard (AJ)~\cite{doersch2022tap} calculated only over these specific track segments (i.e., for all frames $t \geq t_r$).
Therefore this metric measures tracking quality on segments that immediately follow a reappearance after an undetectability period of at least $\dmin$ frames. It can be plotted against $\dmin$ to show how post-reappearance tracking quality behaves as the duration of undetectability increases. To report a summary score across different intervals, we define $\ajrd$ as the mean of $\ajrd^{\dmin}$ over several $\dmin$ values $\dmin \in \{1, 4, 16, 64, 256\}$. This value is calculated \textit{per sample} and averaged over all samples in a dataset. We calculate this metric for PointOdyssey and RoboTAP as both contain long sequences and many reappearance events.

\begin{table*}[h!]
\caption{\textbf{Impact of Augmentations.} Data augmentation while finetuning TAPNext on PointOdyssey (PO) impacts generalizability. While less augmentations lead to higher results on PO, using roll and aspect ratio augmentations lead to overall better results across other datasets.}
\label{tab:augmentation_ablations}
\resizebox{\linewidth}{!}{
  \scriptsize %
  \begin{tabular}{r *{2}{c} *{4}{c} *{3}{c} *{4}{c} *{3}{c}}
  \toprule
    \textbf{\#} & \multicolumn{2}{c}{\textbf{Aug.}} & \multicolumn{4}{c}{\textbf{PointOdyssey}} & \multicolumn{3}{c}{\textbf{DAVIS}} & \multicolumn{4}{c}{\textbf{RoboTAP}} & \multicolumn{3}{c}{\textbf{Kinetics}}\\
    \cmidrule(lr){2-3} \cmidrule(lr){4-7} \cmidrule(lr){8-10} \cmidrule(lr){11-14} \cmidrule(lr){15-17}
     & Roll & Aspect & $\delta^{avg}$ $\uparrow$ & Survival $\uparrow$ & MTE $\downarrow$ & $\ajrd$ $\uparrow$ & AJ $\uparrow$ & $\delta^{avg}$ $\uparrow$ & OA $\uparrow$ & AJ $\uparrow$ & $\delta^{avg}$ $\uparrow$ & OA $\uparrow$ & $\ajrd$ $\uparrow$ & AJ $\uparrow$ & $\delta^{avg}$ $\uparrow$ & OA $\uparrow$\\
    \midrule
1 &$\checkmark$&$\checkmark$ & 49.3 & 63.5 & 17.4 & 20.4 & \textbf{66.0} & \textbf{79.5} & \textbf{91.9} & \underline{59.8} & \underline{74.2} & \underline{88.4} & \underline{51.1} & \underline{53.6} & \underline{68.3} & \underline{88.5}\\
    \midrule
2 &  &  & \underline{51.4} & \underline{68.3} & \textbf{15.6} & \textbf{22.8} & 64.7 & 78.4 & \underline{91.6} & 59.3 & 73.7 & 88.0 & 50.7 & 53.0 & 68.0 & 88.2 \\
3 &  &$\checkmark$ & \textbf{52.0} & \textbf{68.6} & 16.4 & \underline{22.3} & \underline{65.3} & \underline{79.1} & 91.4 & 59.2 & 73.9 & 88.0 & 50.0 & 52.6 & 67.8 & 87.8 \\
4 &$\checkmark$&  & 49.7 & 65.0 & \underline{15.8} & 20.7 & 64.7 & 78.8 & 91.4 & \textbf{60.4} & \textbf{74.7} & \textbf{88.7} & \textbf{52.0} & \textbf{53.7} & \textbf{68.4} & \textbf{88.6} \\
\bottomrule
\end{tabular}
}
\end{table*}

\begin{table*}[h!]
\caption{\textbf{Multi-Dataset Finetuning.} Incorporating multiple datasets in the finetuning helps generalizability while also improving the results on PointOdyssey (PO) compared to training on PO alone.}
\label{tab:dataset_ablations}
\resizebox{\linewidth}{!}{
  \scriptsize %
  \begin{tabular}{r
*{3}{c} *{4}{c} *{3}{c} *{4}{c} *{3}{c}}
  \toprule
    \textbf{\#} & \multicolumn{3}{c}{\textbf{Dataset}} & \multicolumn{4}{c}{\textbf{PointOdyssey}} & \multicolumn{3}{c}{\textbf{DAVIS}} & \multicolumn{4}{c}{\textbf{RoboTAP}} & \multicolumn{3}{c}{\textbf{Kinetics}} \\
    \cmidrule(lr){2-4} \cmidrule(lr){5-8} \cmidrule(lr){9-11} \cmidrule(lr){12-15} \cmidrule(lr){16-18}
     & PO & Kub1024 & \internalds & $\delta^{avg}$ $\uparrow$ & Survival $\uparrow$ & MTE $\downarrow$ & $\ajrd$ $\uparrow$ & AJ $\uparrow$ & $\delta^{avg}$ $\uparrow$ & OA $\uparrow$ & AJ $\uparrow$ & $\delta^{avg}$ $\uparrow$ & OA $\uparrow$ & $\ajrd$ $\uparrow$ & AJ $\uparrow$ & $\delta^{avg}$ $\uparrow$ & OA $\uparrow$ \\
    \midrule
1 & $\checkmark$ & $\checkmark$ & $\checkmark$ & \textbf{52.6} & \underline{67.9} & \textbf{13.4} & \textbf{23.3} & \textbf{66.6} & \textbf{79.9} & \textbf{92.1} & \textbf{61.1} & \underline{75.2} & \textbf{89.6} & \underline{52.4} & \textbf{54.4} & \textbf{68.7} & \textbf{89.0}\\
2 & $\checkmark$ & $\checkmark$ &  & \underline{51.8} & \textbf{71.0} & \underline{14.4} & \textbf{23.3} & 65.6 & 79.0 & \underline{92.0} & \textbf{61.1} & \textbf{75.4} & \underline{89.0} & \textbf{52.6} & \underline{53.9} & \underline{68.4} & \underline{88.7}\\
3 & $\checkmark$ &  &  & 49.3 & 63.5 & 17.4 & \underline{20.4} & \underline{66.0} & \underline{79.5} & 91.9 & \underline{59.8} & 74.2 & 88.4 & 51.1 & 53.6 & 68.3 & 88.5\\
\bottomrule
\end{tabular}
}
\vspace{-0.5cm}
\end{table*}

 \subsection{Experimental Setup}

We conduct extensive experiments to evaluate the effectiveness of our proposed training adaptations. All models were fine-tuned based on the publicly available BootsTAPNext-B checkpoint. We performed fine-tuning for 20,000 steps per dataset utilizing the AdamW optimizer~\cite{loshchilov2017decoupled} with a cosine annealing learning rate schedule~\cite{loshchilov2016sgdr}. The peak learning rate was set to $2\times10^{-5}$, preceded by a 100-step linear warmup. For coordinate prediction, we implemented a loss re-weighting strategy where contributions from points that are occluded but remain within the frame are down-weighted by a factor of $0.2$. For fine-tuning, we use 8 NVIDIA H100 GPUs with an effective batch size of 1 while the evaluations were performed on a single H100. 

\paragraph{Training Datasets} In this work we used a combination of video and image datasets. We utilized three primary video datasets providing ground-truth point tracks for both dynamic objects and static background regions: 1) The training split of PointOdyssey v1.2~\cite{Zheng_2023_ICCV}, which comprises 131 videos featuring diverse object types and complex, long-term camera trajectories. 2) A newly generated synthetic long-sequence Kubric dataset (Kubric-1024), consisting of 10,000 videos, each 1024 frames in length. This dataset models dynamic scenarios, including falling and randomly jumping objects under various camera motions. 3) A novel synthetic dataset, also newly generated, containing 10,000 videos of 240 frames each, depicting dynamic indoor scenes with human motion and camera movement (\internalds). For our initial long-sequence experiments, we also employed 10,000 images from OpenImages v7~\cite{kuznetsova2020open}. These images were used to artificially synthesize long-sequence videos of 1024 frames by applying smooth in-plane homography transformations and the aforementioned roll augmentation to a single static image. Ground-truth point trajectories were derived from randomly sampled points on the image.

\paragraph{Evaluation Datasets} We evaluate our method on both real-world and synthetic video datasets. We use three \textit{real} benchmarks: TAP-Vid-DAVIS~\cite{doersch2022tap}, which contains 30 videos (34-104 frames) from the DAVIS 2017 validation set~\cite{pont20172017}; TAP-Vid-Kinetics~\cite{doersch2022tap}, consisting of 1,189 videos (250 frames each) from the Kinetics-700 validation set~\cite{carreira2017quo}; and RoboTAP~\cite{vecerik2024robotap}, with 265 videos (average 271.9 frames). To quantify long-term tracking performance, we use the \textit{synthetic} PointOdyssey v1.2 dataset~\cite{Zheng_2023_ICCV}, featuring 13 videos with 1,149-4,325 frames in its test split. We also report results on TAP-Vid-RGB-Stacking~\cite{doersch2022tap}, a second synthetic benchmark consisting of 50 videos of 250 frames each.
All images are resized to $256\times256$ for evaluation unless otherwise stated.

\subsection{Evaluations}
The efficacy of recurrent architectures in point tracking is limited by state degradation over long sequences, hindering the application in long-term tracking scenarios. To quantify this limitation, we first analyze the memory capability of BootsTAPNext. We conducted a controlled experiment by tracking points on a static image sequence: query points are initialized on the first frame, and since the scene is static, ground-truth points remain stationary. As illustrated in \cref{fig:longevity}, BootsTAPNext's baseline performance degrades significantly after approximately 150 frames even when the same static frame is fed as input at every time step, indicating a loss of information about initial query locations in its recurrent state. We hypothesized that training on longer video sequences could mitigate this effect. Indeed, fine-tuning on \internalds, which consists of 240-frame videos, extended the model's effective memory to 600 frames in the same static-image experiment.

This initial result motivates us to train on even longer sequences of 1024 frames. We first explore synthetically generated long videos from OpenImages, using homography and roll augmentations to simulate motion. Although this approach further improved memory retention, as shown in \cref{fig:longevity}, training exclusively on this synthetic data proved detrimental to the model's overall tracking performance across standard benchmarks, achieving low Average-Jaccards of $25.00$ (DAVIS), $41.00$ (RoboTAP) and $30.76$ (Kinetics). We attribute this to the model overfitting to homographic transformations, compromising its ability to generalize to more complex, real-world motion.

Consequently, we shifted to fine-tuning using the PointOdyssey dataset, picking a random window of 1024 frames for each sample. Fine-tuning TAPNext on PointOdyssey extended its memory capacity significantly further than either \internalds~or synthetic OpenImages data (\cref{fig:longevity}). Furthermore, fine-tuning on PointOdyssey using random crops with a standard 1:1 aspect ratio yielded substantial improvements on the PointOdyssey test set, leading to state-of-the-art results including an enhanced survival rate of $68.31$ (\cref{tab:augmentation_ablations}, row 2), confirming the potential of the TAPNext architecture for long-term tracking. However, this specialized training led to reduced performance on other benchmarks like DAVIS, indicating overfitting to the PointOdyssey training set. 

To improve generalization, we incorporate aggressive roll augmentations and variable aspect ratio crops during fine-tuning. As shown in \cref{tab:augmentation_ablations} (row 1 vs. row 2), these augmentations improve performance across multiple benchmarks (e.g., boosting DAVIS AJ from $64.7\%$ to $66.0\%$) while maintaining a high survival rate of $63.5\%$ on PO. Additionally, roll augmentations significantly improve re-detection capabilities, as shown in \cref{fig:aj_rd_plot}.

Given PointOdyssey's limited size (131 training samples), we further enhance robustness by fine-tuning on a mixture of datasets: PointOdyssey, Kubric-1024, and \internalds. As shown in \cref{tab:dataset_ablations}, incorporating multiple datasets improves generalization and boosts PointOdyssey performance beyond training on PointOdyssey alone. While mixing PointOdyssey and Kubric-1024 yields the highest PO survival rate ($71.0\%$), mixing all three datasets (\cref{tab:dataset_ablations}, row 1) provides the best overall performance across all benchmarks and constitutes our final model, \methodname{}.
Our fine-tuning strategy on TAPNext achieves a new state of the art in long-term tracking. As shown in \cref{tab:benchmark_results}, \methodname{} outperforms previous methods on PointOdyssey ($\delta^{avg}$ $52.6$, Survival $67.9\%$, $\ajrd$ $23.3$), DAVIS (AJ $66.6\%$, $\delta^{avg}$ $79.9$, OA $92.1\%$), and RGB-Stacking (AJ $73.4\%$, OA $95.1\%$). Notably, it achieves the highest mean $\delta^{avg}$ ($72.2$) across all benchmarks, despite using a lower input resolution ($256\times256$) than competitors and having the lowest latency. This contrasts sharply with BootsTAPNext-B, which struggles on PointOdyssey (Survival $13.0\%$), demonstrating our method successfully addresses its long-sequence limitations. Our model also achieves the best $\ajrd$ on PointOdyssey and RoboTAP, highlighting its superior re-detection capabilities. 
Finally, we evaluate a high-resolution version of our model. The input images are directly resized to $512\times512$ resolution without prior resizing to $256 \times 256$ and the TAP metrics are calculated on the $256 \times 256$ scale. The high-resolution model achieves $\delta^{avg}$ $52.2$, AJ $33.5\%$, Survival $69.1\%$ on PointOdyssey and $\delta^{avg}$ $80.0$, AJ $66.9\%$, OA $92.5\%$ on Davis.
In addition to the advantages discussed above, we compare inference speed among different online trackers in \cref{tab:speed}, demonstrating the low latency and high FPS of \methodname.
Qualitative comparisons are shown in \cref{fig:qual_ex} and \cref{fig:clock}, highlighting that our method is the only one successfully tracking the challenging re-detection and long-term scenarios.

\section{Conclusion}
This work addressed key limitations of current Tracking-Any-Point (TAP) models: their degraded performance on long sequences and their inability to re-detect points following occlusion or points re-entering the frame. We presented \methodname, a model that re-purposes the efficient TAPNext architecture to show that these failures are not architectural but are a result of insufficient training strategies and datasets.
We successfully scaled the model to long sequences by fine-tuning on 1024-frame videos, which we enabled through multi-GPU sequence parallelism. To specifically target re-detection failures, we introduced tailored geometric augmentations and evaluated their effectiveness. Furthermore, we identified re-detection as a blind spot in current evaluations and proposed a new metric called Re-Detection Average Jaccard ($\ajrd$), to rigorously measure post-reappearance tracking quality. We encourage the community to further work on remaining limitations of TAPNext++ and other point tracking approaches, such as solving hour-long point tracking with constant memory, tracking occluders and modeling motion ambiguities.
Our resulting model, \methodname, achieves state-of-the-art performance on multiple point tracking benchmarks while retaining the low memory and compute footprint required for real-world applications.

{
    \small
    \bibliographystyle{ieeenat_fullname}
    \bibliography{main}
}

\setcounter{page}{1}
\setcounter{section}{0}
\renewcommand{\thesection}{\Alph{section}}
\clearpage
\setcounter{page}{1}
\maketitlesupplementary

\newcommand{\R}{\mathbb{R}}
\newcommand{\defeq}{\mathrel{\overset{\makebox[0pt]{\mbox{\tiny def}}}{=}}}

\section{Frequently Asked Questions}

\noindent\textbf{Q: Is the improved performance on PointOdyssey simply due to training on the PointOdyssey dataset?}

\noindent To verify generalization, we trained a variant of \methodname~exclusively on Kubric-1024 and \internalds, excluding PointOdyssey from training entirely. As shown in \cref{tab:training_without_po}, this zero-shot variant still outperforms BootsTAPNext-B by $38.3$ percentage points in survival rate and exceeds Track-On by $3.8$ points, demonstrating that the long-sequence training recipe generalizes to out-of-domain videos beyond the PointOdyssey distribution.

\begin{table}[h]
\caption{\textbf{Training without PointOdyssey.} Evaluation results on PointOdyssey (PO) with and without including PO in the training mix.}
\label{tab:training_without_po}
\centering
\resizebox{\columnwidth}{!}{
    \begin{tabular}{l c c c}
    \toprule
    Method & $\delta^{avg}$ $\uparrow$ & Survival $\uparrow$ & MTE $\downarrow$ \\
    \midrule
    CoTracker3 & 44.5 & 56.3 & 20.7 \\
    Track-On & 35.4 & 47.5 & 33.5 \\
    BootsTAPNext-B (Baseline) & 9.9 & 13.0 & 92.1 \\
    Ours (w/ PO) & 52.6 & 67.9 & 13.4 \\
    \textbf{Ours (w/o PO)} & 37.3 & 51.3 & 26.4 \\
    \bottomrule
    \end{tabular}
}
\end{table}

\noindent\textbf{Q: Does the roll augmentation give \methodname~an unfair advantage on $\ajrd$?}

\noindent We isolate the effect of roll augmentation in \cref{tab:augmentation_ablations}. While it benefits RoboTAP and Kinetics, it slightly reduces scores on PointOdyssey and DAVIS. Crucially, even \textit{without} roll augmentation, our model remains state-of-the-art on DAVIS ($\text{AJ}=65.3$) and PointOdyssey ($\delta^{avg}=52.0$, Survival$=68.6$, MTE$=16.4$, $\ajrd=22.3$). The roll augmentation is a principled, task-specific training technique that any method can equally adopt; it is not a post-hoc advantage restricted to \methodname.

\noindent\textbf{Q: Does \methodname~claim to solve infinite-length point tracking?}

\noindent No. We explicitly target long but finite sequences and do not claim that performance degradation is fully eliminated at arbitrary sequence lengths. Our contribution is to demonstrate that such degradation is \textit{not} an inherent architectural limitation of \textit{linear} recurrent units---in contrast to standard non-linear recurrent architectures that suffer from vanishing gradients. By extending the effective tracking range from $\sim$150 frames to over 4000 frames without any architectural change, we provide new insights into the practical scalability of linear recurrent units for point tracking.

\noindent\textbf{Q: Is \methodname~merely a ``bag of tricks'' with limited technical novelty?}

\noindent We argue that system-algorithmic co-design is a fundamental research contribution, particularly in the context of scalable training. Concretely: (1)~the distributed parallel scan for SSM blocks is not a standard engineering detail but the technical enabler that makes end-to-end training on 1024-frame sequences feasible on multiple GPUs---without it, na\"ive sequence parallelism would require $\mathcal{O}(N)$ sequential communication rounds instead of $\mathcal{O}(\log N)$; (2)~the roll augmentation is a principled, task-specific design directly addressing the well-known failure mode of points exiting and re-entering the frame; (3)~$\ajrd$ fills a concrete gap in the evaluation literature by directly measuring post-reappearance tracking quality, which no prior metric captures.

\noindent\textbf{Q: How does Kubric-1024 differ from prior Kubric datasets?}

\noindent Kubric-1024 required a fundamental redesign to sustain scene dynamics over 1024 frames. In prior Kubric variants, object motion dissipates quickly due to passive physics. We address this by introducing ``velocity bumps''---random linear and angular impulses applied whenever an object's speed drops below a threshold, with a slight bias toward the scene origin to keep objects within the camera's field of view throughout the full sequence. We additionally apply sinusoidal noise to both the camera position and look-at point to simulate natural camera jitter, preventing the model from overfitting to the unnaturally smooth trajectories present in prior synthetic datasets.

\noindent\textbf{Q: Why does the high-resolution ($512\times512$) variant sometimes underperform the $256\times256$ model in \cref{tab:benchmark_results}?}

\noindent To ensure a fair comparison, the high-resolution variant was fine-tuned using the same training recipe as the $256\times256$ model rather than a separately tuned configuration. The optimal hyperparameters for higher-resolution fine-tuning---including learning rate, schedule length, and batch composition---likely differ and were not optimized in this work. The results in \cref{tab:benchmark_results} should therefore be treated as a conservative lower bound on the potential of the $512\times512$ model.

\section{Long-Sequence Training with Sequence Parallelism}

\label{supplementary_sec:long_sequence_parallelism_supp}
Training TAPNext on long sequences ($T=1024$ frames or more) requires fitting large computation graphs and intermediate activations into memory, which exceeds single-device limits. To enable end-to-end training on such sequences, we adopt Sequence Parallelism (SP). The input sequence is partitioned into $N$ chunks along the temporal dimension, with each chunk processed by a dedicated GPU. This strategy requires communicating $\mathcal{O}(T^2)$ bits for attention-based models via context parallelism. At the same time the linear recurrent nature of the State Space Models (SSMs) allows performing forward and backward propagation with only $\mathcal{O}(\log T)$ sequential and $\mathcal{O}(T \log T)$ total communication bits as we show below.

The Real-Gated Linear Recurrent Unit (RG-LRU) used in TAPNext, like other SSMs, is defined by a linear recurrence relation of the form:
\begin{equation}
    h_t = a_t \odot h_{t-1} + x_t,
\end{equation}
where $h_t, a_t, x_t \in \R^D$ are the hidden state, recurrence parameters, and input at time $t$, respectively, and $\odot$ denotes element-wise multiplication. This recurrence can be expressed as an associative binary operator. First, if $h_1 = a_1 \odot h_0 + x_1$ and $h_2 = a_2 \odot h_1 + x_2$, then $h_2 = (a_2 \odot a_1) \odot h_0 + (a_2 \odot x_1 + x_2)$. This allows us to define an associative binary operator $\oplus$:
\begin{equation}
    (a_2, x_2) \oplus (a_1, x_1) = (a_2 \odot a_1, a_2 \odot x_1 + x_2).
    \label{eq:assoc_op}
\end{equation}
Because this operator is associative, we can compute the recurrence using a parallel scan algorithm, rather than a sequential loop. We leverage this property to implement a \textit{distributed parallel scan} for RG-LRU during the forward and backward passes, enabling efficient training on distributed sequence chunks. Our distributed scan operates in three phases:

\begin{enumerate}
    \item \textbf{Local Scan:} Each GPU $j \in \{0, \dots, N-1\}$, holding sequence chunk $\{(a^j_t, x^j_t)\}_{t=0}^k$, computes a local parallel scan assuming its initial hidden state $h^j_{\text{in}}$ is zero. This produces a preliminary local output $y'^j$ and a chunk summary $S_j = (\alpha_j, \chi_j)$, where $\alpha_j = a^j_k \odot \dots \odot a^j_0$ and $\chi_j = y'^j_k$. This summary $S_j$ represents the affine transformation of the chunk: $h^j_{\text{out}} = \alpha_j \odot h^j_{\text{in}} + \chi_j$. This phase is executed in parallel across all $N$ GPUs.

    \item \textbf{Cross-GPU Prefix Scan:} To find the correct input state $h^j_{\text{in}}$ for each chunk $j>0$, we must compute $h^j_{\text{in}} = h^{j-1}_{\text{out}}$. This requires composing the transformations of all preceding chunks: $h^j_{\text{in}} = (\alpha_{P_{j-1}} \odot h_{\text{start}}) + \chi_{P_{j-1}}$, where $P_{j-1} = S_{j-1} \oplus S_{j-2} \oplus \dots \oplus S_0 = (\alpha_{P_{j-1}}, \chi_{P_{j-1}})$ and $h_{\text{start}}$ is the initial state of the entire sequence (typically zero). Computing all prefix compositions $\{P_0, P_1, \dots, P_{N-1}\}$ is a standard prefix scan problem over the set of summaries $\{S_0, S_1, \dots, S_{N-1}\}$ using the operator $\oplus$. This is solved efficiently across $N$ GPUs using tree-based algorithms (e.g., recursive doubling) with $O(\log N)$ communication rounds, rather than $O(N)$ rounds required by sequential propagation.

    \item \textbf{Local Correction:} Once each GPU $j$ receives its correct initial state $h^j_{\text{in}}$ from Phase 2, it corrects its preliminary output $y'^j$ by adding the contribution from $h^j_{\text{in}}$: $y^j_t = y'^j_t + c^j_t \odot h^j_{\text{in}}$, where $c^j_t = a^j_t \odot a^j_{t-1} \odot \dots \odot a^j_0$. This phase is also performed in parallel across all GPUs.
\end{enumerate}

The backward pass follows a similar distributed scan pattern to propagate gradients efficiently across GPUs.

In addition to that, we also implement sequence parallelism for temporal causal convolution layers. Unlike the linear recurrence, each step of the sequence may only depend on a constant number of past steps in forward pass for temporal convolution (or future steps during backward pass). Therefore, we only need to do steps 1 and 2 and for the latter we can do one linear chain of communications in parallel. 

This method allows TAPNext to be trained end-to-end on sequences significantly longer than those trainable on a single device, with only a logarithmic increase in communication overhead with respect to the number of GPUs.

\section{Dynamic Model Depth}
Prior work\cite{Zholus_2025_ICCV} has shown that TAPNext-like architectures can be trained using a layer-wise supervision strategy, meaning that each layer is trained using its own loss. As a result, the model supports dynamic inference depth: it can be executed with fewer layers at test time without requiring retraining. Although the original architecture uses twelve layers, we observed that the predictions produced after only eight layers are already comparable in quality to those obtained after all twelve. This allows one to reduce memory consumption and further increase the achievable inference speed (FPS). Note that in this work, all experiments were executed using the full twelve layers.

\begin{figure}[t]
    \centering
    \includegraphics[width=1\columnwidth]{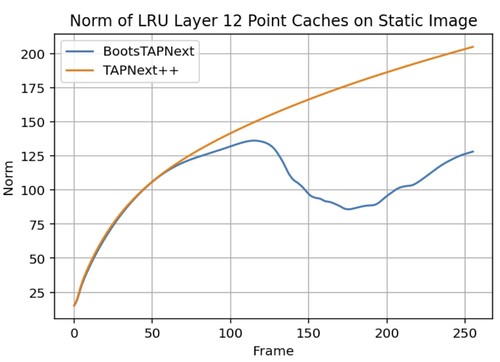}
    \caption{\textbf{LRU State Norm Growth on re-feeding a Single Image.} BootsTAPNext shows unstable LRU state after 100 frames. \methodname~has a monotonically growing state.}
    \label{fig:lru_state}
\end{figure}

\begin{figure}[t]
    \centering
    \includegraphics[width=1\columnwidth]{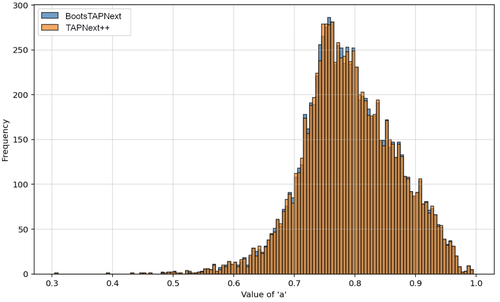}
    \caption{\textbf{Eigenvalue Distribution.} Distribution of eigenvalues $a$. Almost no change can be observed between BootsTAPNext and \methodname.}
    \label{fig:eigenvalue_dist}
\end{figure}

\section{LRU State analysis}
We analyze the temporal behavior of the LRU state associated with the point-token stream in the twelfth layer. To isolate the intrinsic LRU dynamics from changes caused by varying inputs, we repeatedly feed the same static image into the network and track the norm of the resulting cache state over time. As shown in \cref{fig:lru_state}, the original BootsTAPNext exhibits irregular and unstable cache growth after approximately 100 frames, whereas TAPNext++ produces a smooth and nearly monotonic evolution of the cache norm. Also note that both network show equal state growth in the beginning of the video. Importantly, even with a constant input image, changes in the LRU state are expected: the LRU update rule operates directly on the current activation and does not explicitly enforce invariance over repeated inputs.

To better understand this behavior, we further inspect the distribution of the learned eigenvalues $a$ that parametrize the LRU update (cf. \cref{fig:eigenvalue_dist}). Surprisingly, we do not observe a pronounced shift toward eigenvalues near one in \methodname, which would correspond to stronger long-term memory retention. This suggests that the improved stability of \methodname~does not arise from extending the effective memory horizon of the LRU block. Instead, the model may have learned to regulate or compensate for long-sequence LRU dynamics elsewhere in the architecture—potentially through more stable interactions between point- and patch-token pathways or through modifications in downstream attention layers.

\end{document}